\documentclass{article}

\PassOptionsToPackage{round, authoryear}{natbib}
\usepackage[dblblindworkshop,final]{neurips_2026}
\makeatletter
\renewcommand{\@noticestring}{Accepted at the AI for Science workshop (NeurIPS 2026).}
\makeatother

\usepackage[utf8]{inputenc}
\usepackage[T1]{fontenc}
\usepackage[hidelinks]{hyperref}
\usepackage{url}
\usepackage{booktabs}
\usepackage{amsmath}
\usepackage{amssymb}
\usepackage{amsfonts}
\usepackage{nicefrac}
\usepackage{microtype}
\usepackage{xcolor}
\usepackage{graphicx}
\usepackage{multirow}
\usepackage{caption}
\usepackage{tikz}
\usetikzlibrary{arrows.meta,positioning}

\graphicspath{{figures/}}

\newcommand{\Ghat}{\hat{G}_\theta}
\newcommand{\Gex}{G_{\mathrm{exact}}}
\newcommand{\Gsch}{G_{\mathrm{scheme}}}
\newcommand{\Ddiss}{\Delta_{\mathrm{diss}}}
\newcommand{\Ddisp}{\Delta_{\mathrm{disp}}}
\newcommand{\Dtwin}{\Delta_{\mathrm{twin}}}
\newcommand{\Dscheme}{D_{\mathrm{scheme}}}
\newcommand{\Dexact}{D_{\mathrm{exact}}}

\title{The Symbol of the Surrogate:\\Measuring Numerical Provenance in Neural PDE Solvers}

\author{%
  Ridham Patel \\
  Computer Science \& Engineering \\
  Indian Institute of Technology Gandhinagar \\
  Gandhinagar, India \\
  \texttt{ridham.patel@iitgn.ac.in}
}

\begin{document}

\maketitle

\begin{abstract}
Neural PDE surrogates are trained on numerical solver outputs that contain both physical evolution and solver-specific discretization errors. Because surrogates are also evaluated against held-out trajectories from the same solver, standard benchmarks cannot distinguish fidelity to the exact evolution from imitation of the numerical scheme. We introduce an empirical Fourier-symbol diagnostic that probes a trained surrogate's linearized one-step operator with individual Fourier modes and compares it with both exact-evolution and training-scheme references. To address architectural spectral bias, we train identical networks on schemes with orthogonal dissipative and dispersive signatures and compare their learned operators. In linear advection, the learned surrogates reproduce the training schemes' amplitude and phase errors, with the twin-scheme difference reaching more than 99.8\% of the analytically predicted full-imitation ceiling. The same behavior occurs for a non-local Fourier neural operator and at the operator level for nonlinear Burgers dynamics. These results show that agreement with solver-generated test data does not by itself establish fidelity to the exact evolution. Fourier-symbol measurements provide a direct diagnostic of numerical provenance.
\end{abstract}

\section{Introduction}

Neural surrogates for partial differential equations (PDEs) are often trained on trajectories
produced by a numerical solver and, in the prevailing protocol, evaluated against held-out
trajectories from the same solver. Benchmarks are constructed this way as a matter of
course: PDEBench and PDEArena distribute solver-generated trajectories as targets, and
standard neural-operator and autoregressive simulator studies evaluate against numerical
rollouts \citep{pdebench,pdearena,fno,mppde,gns}.

Solver output is not the physical evolution. Every consistent scheme introduces a
discretization error whose wavenumber dependence is classically characterized: first-order
upwind adds artificial diffusion that damps high wavenumbers, while centered and
Crank--Nicolson discretizations of advection conserve amplitude but propagate high
wavenumbers at the wrong speed \citep{leveque}. A surrogate trained on such data is
therefore exposed to two superimposed signals, the physical evolution and the solver's
numerical signature, and the training objective does not separate them.

This creates an ambiguity that the standard protocol cannot resolve. If the test targets are
generated by the same scheme as the training targets, then the benchmark rewards agreement
with the scheme, and a surrogate that has reproduced the scheme's numerical signature can
score as well as, or better than, a surrogate that has learned the underlying evolution. A low
error against held-out solver trajectories therefore does not establish fidelity to the exact
evolution; it establishes fidelity to the solver, of which fidelity to the exact evolution is
only one possible cause. This ambiguity is adjacent to prior work on solver-aware learning,
low-fidelity supervision and learned correction of numerical solver errors
\citep{solverclosures,emulatorsuperiority,solverintheloop,multifidelitydeeponet}.

The closest spectral diagnostic does not close this gap. \citet{spectralaudit} measure how
far a learned operator sits from the exact operator, which quantifies error but does not
attribute it: the two candidate targets, the exact evolution and the training scheme, are
never compared side by side. What is missing is an instrument that reports which of the two a
learned operator is near.

We use the surrogate's \emph{empirical Fourier symbol} as that instrument. Following the
single-mode probing primitive of \citet{spectralaudit}, we apply the trained surrogate to one
Fourier mode and read the corresponding output coefficient in one forward pass, with no
gradients. The same object exists in closed form for both references: $\Gex(k)$ for the exact
evolution and $\Gsch(k)$, the von~Neumann amplification factor, for the scheme. Provenance
then becomes a projection: a scalar $\alpha(k)$ that is $1$ when the learned symbol coincides
with the scheme and $0$ when it coincides with the exact reference.

One confound stands in the way of reading $\alpha$ directly. Neural networks are biased
toward low frequencies for architectural reasons \citep{rahaman,tancik,spectralbiasop}, and
that bias damps high wavenumbers, which is the same qualitative effect a dissipative scheme
has. A
single surrogate trained on a dissipative scheme would report $\alpha > 0$ even if it had
learned no scheme-specific behaviour at all.

Our identification strategy removes this confound by construction. We train identical
networks on two schemes whose errors are orthogonal (upwind, whose error is almost purely
dissipative, and Crank--Nicolson, whose error is purely dispersive) on one shared mesh and
time step, with every other experimental factor held fixed \citep{leveque}. Under those conditions, the
scheme-independent component of the architectural contribution is suppressed in the
difference of the two learned symbols. Target-dependent approximation and optimization
effects need not cancel, while the scheme contributions separate onto
orthogonal axes. Spectral bias alone predicts a twin-scheme difference of zero; scheme
imitation predicts a specific, analytically computable value.

The measurements that follow are consistent with scheme imitation and not with the
alternatives. On linear advection the twin-scheme difference recovers over $99.8\%$ of its
full-imitation ceiling on both axes; the same holds for a non-local Fourier neural operator
that demonstrably fits the exact evolution when trained on exact data; on nonlinear Burgers
the learned operator remains aligned with its training scheme by $5$--$27\times$ on its own states,
including under common-input evaluation; and fine-tuning on $25$ exact samples moves $\alpha$ to
approximately $0$ without removing the architecture-dependent accuracy floor.

We make three contributions.
\begin{itemize}\itemsep0pt
  \item We turn the empirical Fourier symbol of a trained surrogate, a measurement primitive
    introduced for operator networks by \citet{spectralaudit}, into a provenance test, by
    resolving the symbol into dissipation and dispersion coordinates and projecting it onto
    the axis joining the exact reference to the training scheme.
  \item We introduce twin-scheme differencing, an identification strategy in which two
    identical networks trained on schemes with orthogonal error signatures suppress the
    common architectural contribution, giving a falsifiable prediction that spectral bias
    alone cannot reproduce.
  \item We apply the diagnostic to standard surrogates and find scheme-consistent symbols for
    a local CNN and a non-local FNO on linear advection, at the operator level on nonlinear
    Burgers with the input distribution held fixed, and we show that provenance and accuracy
    respond on different scales under high-accuracy fine-tuning.
\end{itemize}

\section{Method}

\subsection{Problem setup}
\label{sec:setup}

We consider a periodic PDE on $[0,L)$ discretized on $N$ equispaced points with spacing
$\Delta x = L/N$, and a fixed time step $\Delta t$. Three one-step maps on the discrete state
$u \in \mathbb{R}^N$ are relevant:

\begin{itemize}\itemsep1pt
  \item the \emph{exact reference} $\mathcal{E}$, the true evolution over $\Delta t$ sampled
    on the grid;
  \item the \emph{scheme reference} $\mathcal{S}^{(s)}$, one step of numerical scheme $s$;
  \item the \emph{learned operator} $f_\theta$, a network trained on pairs
    $(u, \mathcal{S}^{(s)}u)$ generated by scheme $s$.
\end{itemize}

Training data are generated by $\mathcal{S}^{(s)}$ and standard evaluation compares
$f_\theta$ to $\mathcal{S}^{(s)}$ on held-out inputs. The question we ask of $f_\theta$ is
different: is the operator it learned closer to $\mathcal{E}$ or to $\mathcal{S}^{(s)}$, and
by how much? Appendix~\ref{app:pipeline} gives a schematic of the measurement pipeline.

\subsection{The empirical Fourier symbol}
\label{sec:symbol}

On a periodic grid, a linear one-step map that commutes with translations acts on each
discrete Fourier mode by multiplication by a complex number: its \emph{symbol}
\citep{leveque,spectralaudit}. We measure
that number for a trained surrogate directly. For each resolved wavenumber
$k \in \{1,\dots,N/2-1\}$ we build the probe field
\begin{equation}
  u^{(k)}_j \;=\; \varepsilon \cos(2\pi k x_j / L), \qquad j = 0,\dots,N-1 ,
\end{equation}
apply the surrogate's one-step map once, take the discrete Fourier transform of the output,
and divide its $k$-th coefficient by the $k$-th coefficient of the input:
\begin{equation}
  \Ghat(k) \;=\;
  \frac{\big[\mathcal{F} f_\theta(u^{(k)})\big]_k}{\big[\mathcal{F} u^{(k)}\big]_k},
  \qquad \varepsilon \to 0 .
  \label{eq:symbol}
\end{equation}
This costs one forward pass per wavenumber and requires no gradients and no access to model
internals. We use $\varepsilon = 10^{-2}$ throughout.

Equation~\eqref{eq:symbol} is a symbol only if the probe stays in the linear regime, so we
certify rather than assume it. Sweeping $\varepsilon \in [10^{-4}, 1]$, the recovered
$\Ghat$ is $\varepsilon$-independent to $1.0\times10^{-2}$ in relative terms across
$\varepsilon \in [10^{-4}, 10^{-1}]$ and departs only at $\varepsilon \sim 1$, where the GELU
nonlinearity \citep{gelu} is no longer probed locally (Appendix~\ref{app:eps}). Applied to the numerical
solvers themselves, the same procedure returns their analytic symbols to $2\times10^{-16}$,
which fixes the zero of the instrument.

For nonlinear problems, the one-step map is not generally a Fourier multiplier. 
Around a
nonconstant base state $\bar u$, the Jacobian $Df_\theta|_{\bar u}$ need not commute with
translations, so a single Fourier input can generate responses in other modes. In this case
\eqref{eq:symbol} should be read as the same-mode, diagonal Fourier response
$\langle e_k, Df_\theta|_{\bar u} e_k\rangle/\langle e_k,e_k\rangle$, not as a full scalar
symbol. We take $\bar u$ from the solver's own forced steady states, the distribution the
surrogate was trained on. This matches the frozen-coefficient viewpoint behind linearized
von~Neumann analysis of nonlinear schemes \citep{leveque}: both operators are queried about
the same occupied state rather than about an unrelated distribution.

\subsection{The two reference symbols}
\label{sec:refs}

Linear advection $u_t + c u_x = 0$ makes the two references explicit. Write
$\sigma = c\Delta t/\Delta x$ for the Courant number and $\varphi_k = 2\pi k/N$ for the phase
advance per cell of mode $k$. The exact evolution translates each mode without changing its
amplitude,
\begin{equation}
  \Gex(k) \;=\; e^{-i\sigma \varphi_k}, \qquad |\Gex| = 1 ,
\end{equation}
whereas a consistent scheme applies a different multiplier, known in closed form. For the two
schemes used throughout,
\begin{equation}
  G^{\mathrm{up}}_{\mathrm{scheme}}(k) = 1 - 2\sigma\sin^2\!\tfrac{\varphi_k}{2}
    - i\,\sigma \sin\varphi_k ,
  \qquad
  G^{\mathrm{CN}}_{\mathrm{scheme}}(k) = \frac{1 - i a_k/2}{1 + i a_k/2},
  \quad a_k = \sigma \sin\varphi_k ,
  \label{eq:schemes}
\end{equation}
for explicit first-order upwind and Crank--Nicolson with centered differences, respectively
\citep{leveque}.
The two schemes err in complementary ways. Upwind damps: $|G^{\mathrm{up}}_{\mathrm{scheme}}| < 1$,
and at the Courant number used here, $\sigma = 1/2$, it factors as
$G^{\mathrm{up}}_{\mathrm{scheme}} = e^{-i\varphi_k/2}\cos(\varphi_k/2)$, so its phase agrees
with $\Gex$ exactly and its entire error is a modulus deficit. Crank--Nicolson is unitary,
$|G^{\mathrm{CN}}_{\mathrm{scheme}}| = 1$, and its entire error is a phase error. At this
Courant number the two scheme errors therefore separate exactly into a modulus component and
a phase component, which is the property the identification strategy of Sec.~\ref{sec:twin}
exploits; away from $\sigma = 1/2$ the upwind error acquires a phase component as well and the
separation becomes approximate.

\subsection{Dissipation and dispersion coordinates}
\label{sec:coords}

Because $|\Gex| = 1$, multiplying by the complex conjugate $\overline{\Gex}$ rotates the
complex plane so that the
exact reference sits at $1$ and the error $\Ghat - \Gex$ decomposes into a radial and a
tangential part:
\begin{equation}
  \begin{aligned}
    \Ddiss(k) &= \mathrm{Re}\big[(\Ghat - \Gex)\,\overline{\Gex}\big]
      &&\approx\; |\Ghat| - 1 , \\[2pt]
    \Ddisp(k) &= \mathrm{Im}\big[(\Ghat - \Gex)\,\overline{\Gex}\big]
      &&\approx\; \arg\Ghat - \arg\Gex .
  \end{aligned}
  \label{eq:coords}
\end{equation}
The left-hand sides are the quantities we report; the right-hand approximations hold to first
order in the deviation and give them their names. $\Ddiss$ is an amplitude error: negative
means the mode is damped relative to the exact evolution. $\Ddisp$ is a phase error: positive
means the mode is under-rotated, i.e.\ it propagates more slowly than the true wave, which is
the classical lagging-phase error of a centered discretization. Modified-equation theory
places each scheme on one of these axes: upwind, whose leading error term is an even
derivative, is radial; Crank--Nicolson, whose leading error term is odd and which is unitary,
is tangential \citep{leveque}.

\subsection{The provenance coefficient}
\label{sec:alpha}

All the quantities needed to attribute the learned symbol are now defined. In the rotated
frame the exact reference sits at the origin of the error plane and the training scheme sits
at $\Gsch - \Gex$. We report where the learned symbol falls along the line joining them:
\begin{equation}
  \alpha(k) \;=\;
  \frac{\mathrm{Re}\big[(\Ghat - \Gex)\,\overline{(\Gsch - \Gex)}\big]}
       {\big|\Gsch - \Gex\big|^2} .
  \label{eq:alpha}
\end{equation}
This is the least-squares projection of the learned error onto the scheme error direction,
treating $\mathbb{C}$ as $\mathbb{R}^2$; the conjugate is required for the numerator to be a
real inner product. Three values orient the scale:

\begin{itemize}\itemsep1pt
  \item $\alpha = 1$: the learned symbol has the same component along the scheme-error
    direction as the training scheme. This does not rule out an additional component
    orthogonal to that direction.
  \item $\alpha = 0$: the learned symbol is \emph{consistent with the exact reference} in the
    scheme-error direction. It does not by itself certify that the network has learned the
    physics; it says the scheme's error direction is absent from the learned symbol.
  \item $\alpha < 0$: the learned symbol lies beyond the exact reference in the direction
    opposite to the scheme, the regime in which an emulator is more accurate than the solver
    that trained it \citep{emulatorsuperiority}.
\end{itemize}

The corresponding orthogonal residual is
\begin{equation}
  r_\perp(k) = \left|\Ghat(k)-\Gex(k)-\alpha(k)\bigl(\Gsch(k)-\Gex(k)\bigr)\right| .
  \label{eq:perp}
\end{equation}
For the linear experiments, the modulus and phase summaries together with Figure~\ref{fig:linear}
show that this residual is negligible relative to the scheme displacement.

\subsection{Twin-scheme identification}
\label{sec:twin}

A single value of $\alpha$ is not sufficient evidence, because the learned symbol carries an
architectural contribution as well as a scheme contribution. Model the symbol learned from
scheme $s$ as
\begin{equation}
  \Ghat^{(s)}(k) \;=\; \Gex(k) \;+\; \alpha_s(k)\big[\Gsch^{(s)}(k) - \Gex(k)\big]
  \;+\; B(k) \;+\; \eta ,
  \label{eq:decomp}
\end{equation}
where $\alpha_s(k)$ is the scheme-imitation coefficient that \eqref{eq:alpha} is meant to
estimate, $B(k)$ is the architecture-intrinsic spectral bias (the displacement the network
would exhibit from its inductive bias and optimization alone), and $\eta$ collects
seed-to-seed noise. Applying \eqref{eq:alpha} to $\Ghat^{(s)}$ returns $\alpha_s(k)$ plus the
projection of $B(k)$ onto the scheme-error direction, so the estimator is unbiased only when
that projection vanishes. It does not vanish for a dissipative scheme: spectral bias damps
high wavenumbers, so $B(k)$ points radially inward, and so does the upwind error. A network
with $B(k) \neq 0$ that had learned nothing scheme-specific would therefore still report
$\alpha_{\mathrm{up}} > 0$. This is the confound a single-scheme study cannot escape.

Two features of the design break the tie. First, $\Gsch^{\mathrm{CN}} - \Gex$ is tangential
and therefore orthogonal to a radial $B(k)$, so $\alpha_{\mathrm{CN}}$ is not contaminated by
spectral bias. Second, and more robustly, if the architecture, mesh, time step, input
distribution, optimizer, schedule, sample count and seed sequence are held fixed across the
two schemes, then the scheme-independent component of $B(k)$ is suppressed in the difference
of the two learned symbols. Target-dependent approximation and optimization effects need not
cancel, so this is an identification strategy rather than an exact causal decomposition. We
therefore report
\begin{equation}
  \Dtwin(k) \;=\; \big[\Ghat^{\mathrm{up}}(k) - \Ghat^{\mathrm{CN}}(k)\big]\,\overline{\Gex(k)} ,
  \label{eq:twin}
\end{equation}
where $\overline{\Gex(k)}$ is again the complex conjugate, so that the real and imaginary parts
of $\Dtwin$ are the twin difference in the dissipation and dispersion coordinates of
\eqref{eq:coords}. Under $\alpha_{\mathrm{up}} = \alpha_{\mathrm{CN}} = 1$ and
exact cancellation of $B$, \eqref{eq:decomp} gives a \emph{full-imitation ceiling} that is
computable from the analytic symbols alone,
$\Dtwin^{\max} = (\Gsch^{\mathrm{up}} - \Gsch^{\mathrm{CN}})\overline{\Gex}$, which supplies
the scale against which a measured $\Dtwin$ is meaningful.

The geometry gives the design its falsifiable content. Because the upwind error is radial and
the Crank--Nicolson error tangential, scheme imitation displaces the two learned symbols
along orthogonal directions, forming a right angle in the $(\Ddiss, \Ddisp)$ plane (exactly so
while the deviations are small, approximately so at the largest wavenumbers, where the
Crank--Nicolson phase error approaches $1$~radian and its symbol curves along the unit circle),
and produces a $\Dtwin$ with both components nonzero and matching the ceiling. Spectral bias alone predicts
both symbols displaced radially, no tangential separation, and $\Dtwin = 0$ in both
components however large the individual displacements are. Consistency with the exact
reference predicts both symbols at the origin. Appendix~\ref{app:predictions} tabulates these
predictions explicitly.

\section{Experimental setup}
\label{sec:setup_exp}

The identification argument requires the solver to be the only source of variation, so within
each experiment the architecture, initialization seed sequence, optimizer, learning-rate
schedule, batch size, epoch count, sample count, mesh, time step and distribution of input
fields are identical across corpora; only the solver that generated the targets differs.

For linear advection ($u_t + u_x = 0$ on $[0,1]$, $N = 64$, $\sigma = 0.5$) we generate three
corpora of $4000$ input/target pairs from random band-limited fields with approximately flat
spectra: upwind, Crank--Nicolson, and an exact spectral control. Surrogates are a periodic
five-layer CNN and, for the architecture control of Sec.~\ref{sec:fno}, an FNO retaining all
$N/2$ resolved modes \citep{fno}; both are trained with Adam \citep{adam} and cosine annealing
\citep{sgdr} over $400$ epochs and
three seeds. For forced viscous Burgers ($u_t + u u_x = \nu u_{xx} + f$ on $[0,2\pi]$,
$N = 128$, $\nu = 0.04$) we use three method-of-lines RK4 solvers that share mesh, time step,
viscosity and forcing and differ only in the convection discretization: dealiased spectral
(used as the high-fidelity reference), first-order upwind, and second-order centered \citep{leveque,boyd}. Surrogates
learn the $M = 15$ substep map. Full architectures, hyperparameters, data-generation details
and the seed protocol are given in Appendix~\ref{app:details}.

\section{Results}

\subsection{Linear advection: direct evidence of numerical provenance}
\label{sec:linear}

The question this experiment answers is whether the operator learned from solver-generated
data carries the training scheme's numerical signature, and whether that signature can be
distinguished from architectural spectral bias. Three corpora (upwind, Crank--Nicolson, and
an exact control) are used to train identical CNNs, and the symbol of each is read by
single-mode probing.

\begin{table}[t]\centering\small
\caption{Linear advection: the learned symbol matches the training scheme rather than the
exact reference. Values are means over the upper half of the resolved spectrum, $\pm$ s.d.\
over 3 seeds. ``Learned $|\Ghat|$'' is the modulus of the measured symbol (exact reference:
$1$) and ``learned lag'' is $\Ddisp$, the phase error relative to the exact reference (exact
reference: $0$); each is paired with the value the training scheme itself takes, so that the
deviations in the table are read as scheme values reproduced rather than as surrogate errors.
The upwind scheme has exactly zero phase error at $\sigma = 1/2$ (Sec.~\ref{sec:refs}), and
the Crank--Nicolson scheme is exactly unitary. $\alpha$ is the provenance coefficient of
\eqref{eq:alpha}; it is undefined for the exact control, which has no scheme to attribute to.}
\begin{tabular}{lccccc}
\toprule
& \multicolumn{2}{c}{reproduced dissipation} & \multicolumn{2}{c}{reproduced dispersion} & \\
\cmidrule(lr){2-3}\cmidrule(lr){4-5}
model & learned $|\Ghat|$ & scheme $|\Gsch|$ & learned lag & scheme lag & $\alpha$ \\
\midrule
upwind          & $0.395 \pm 0.000$ & 0.395 & $0.003 \pm 0.003$ & $0$ (exact) & $1.000 \pm 0.000$ \\
Crank--Nicolson & $1.000 \pm 0.000$ & 1.000 & $0.824 \pm 0.000$ & 0.824 & $1.000 \pm 0.000$ \\
exact (control) & $0.987 \pm 0.006$ & -- & $0.025 \pm 0.003$ & -- & -- \\
\bottomrule
\end{tabular}
\label{tab:linear}
\end{table}

Table~\ref{tab:linear} reports the result. The upwind-trained surrogate reproduces the upwind
scheme's amplitude deficit to three decimals ($|\Ghat| = 0.395$ against a scheme value of
$0.395$) while leaving the phase almost untouched, which is what the upwind scheme itself
does at this Courant number. The Crank--Nicolson-trained surrogate remains unitary to three
decimals and reproduces the scheme's phase error ($0.824$ against $0.824$). For both,
$\alpha = 1.000 \pm 0.000$: the learned symbols are scheme-consistent, not
exact-reference-consistent. The exact-trained control sits near, but not at, the exact
reference ($|\Ghat| = 0.987$, lag $0.025$), which is the architectural displacement $B(k)$
made visible.

\begin{figure}[t]\centering
\includegraphics[width=\linewidth]{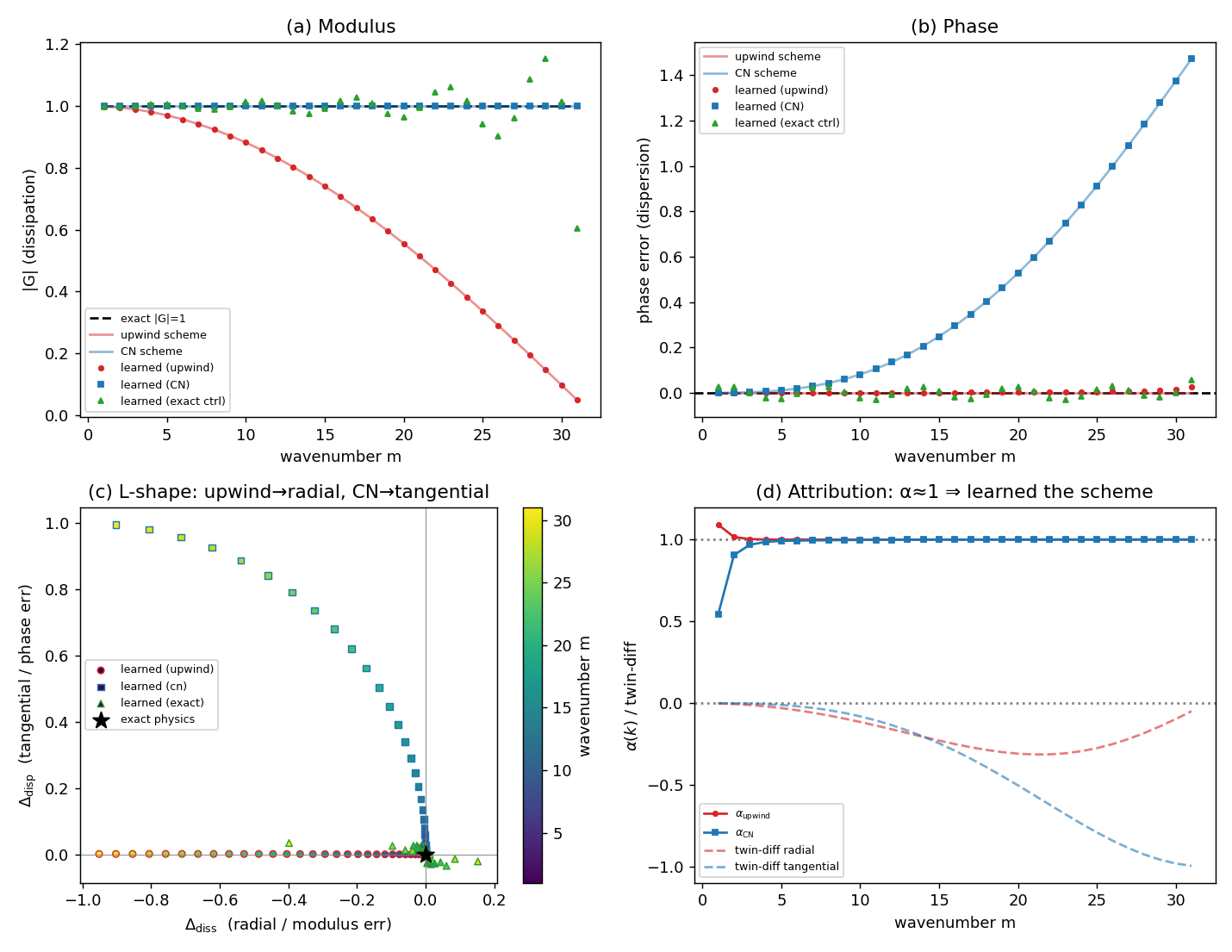}
\caption{Linear advection, per wavenumber $m$ (single seed; Table~\ref{tab:linear} aggregates
over seeds). \textbf{(a)} modulus of the measured symbol against wavenumber: the
upwind-trained surrogate (red dots) lies on the upwind scheme curve (red line) across the
spectrum, while the Crank--Nicolson-trained surrogate (blue squares) stays at $|\Ghat| = 1$
like its unitary scheme; the exact-trained control (green triangles) scatters about $1$ at
high $m$. \textbf{(b)} phase error $\Ddisp$ against wavenumber: only the Crank--Nicolson
surrogate departs from zero, and it tracks the Crank--Nicolson scheme curve. \textbf{(c)} the
two error coordinates plotted against each other, colour-coded by wavenumber: the upwind
surrogate moves along the dissipation axis, the Crank--Nicolson surrogate along the
dispersion axis, and the exact control stays near the origin (star). This orthogonal
separation is the takeaway: each surrogate deviates from the exact reference in the direction
its own training scheme deviates, and the tangential arm is what spectral bias cannot produce.
(The Crank--Nicolson arm bends leftward at the highest wavenumbers, where the phase error
approaches $1$~radian and the first-order reading $\Ddiss \approx |\Ghat| - 1$ of
\eqref{eq:coords} no longer applies.) \textbf{(d)} the provenance coefficient
$\alpha(k) \approx 1$ for both surrogates across the resolved spectrum, and the two components
of the twin-scheme difference.}
\label{fig:linear}
\end{figure}

Figure~\ref{fig:linear}(c) shows the geometry the identification strategy predicts. In the
$(\Ddiss, \Ddisp)$ plane the upwind surrogate lies along the dissipation axis, the
Crank--Nicolson surrogate rises along the dispersion axis, and the exact control sits near the
origin. The tangential arm is the component that the spectral-bias-only model of
\eqref{eq:decomp} cannot produce, since $B(k)$ is scheme-independent and radial.

The twin-scheme difference makes this quantitative. Measured over the upper spectrum,
$\mathrm{Re}\,\Dtwin = -0.2406 \pm 0.0001$ and
$\mathrm{Im}\,\Dtwin = -0.6841 \pm 0.0008$. These numbers are only interpretable against a
scale, and Sec.~\ref{sec:twin} supplies one: the full-imitation ceiling computed from the
analytic symbols is $-0.2406$ radial and $-0.6845$ tangential on this grid. The measurement
therefore recovers $100.0\%$ of the available dissipation signal and $99.9\%$ of the
available dispersion signal. The twin difference is not merely nonzero in the predicted
direction; it is saturated on both axes independently.

\subsection{A non-local architecture: ruling out locality and capacity}
\label{sec:fno}

The locality observation suggests an alternative explanation for Sec.~\ref{sec:linear}: a CNN
represents local finite-difference stencils easily, and the two schemes \emph{are} local
finite-difference stencils, so the agreement might reflect the architecture's hypothesis
class rather than anything about training on solver data. Under that explanation the effect
should weaken or disappear for an architecture that can represent the exact evolution.

We therefore repeat the twin-scheme experiment with an FNO \citep{fno}, which is non-local by
construction and retains all $N/2$ resolved modes. Its spectral convolution has the Fourier
degrees of freedom needed for the exact advection multiplier, and the exact-trained run
serves as the empirical capacity check (Appendix~\ref{app:fno_repr}). Corpora,
optimizer, schedule, sample count and seeds are unchanged.

\begin{table}[t]\centering\small
\caption{Architecture control on linear advection (mean $\pm$ s.d., 3 seeds). Final MSE is
against the corpus the model was trained on; $|\Ghat|$ and lag are the measured symbol
statistics of Table~\ref{tab:linear}; $\alpha$ is the provenance coefficient against the
training scheme. The non-local FNO reproduces its training scheme as strongly as the local
CNN. The exact-trained control (third row of each block) separates the architectures: the CNN
cannot fit the non-local exact evolution, whereas the FNO fits it to $3.6\times10^{-8}$ and
lands at the exact reference, which is what makes the FNO's behaviour on scheme corpora a
consequence of the training data rather than of its hypothesis class.}
\begin{tabular}{llcccc}
\toprule
& corpus & final MSE & mean $|\Ghat|$ & mean lag & $\alpha$ \\
\midrule
\multirow{3}{*}{CNN (local)}
 & upwind & $1.2\times10^{-7}$ & $0.395 \pm 0.000$ & $0.003 \pm 0.003$ & $1.000 \pm 0.000$ \\
 & Crank--Nicolson & $3.6\times10^{-8}$ & $1.000 \pm 0.000$ & $0.824 \pm 0.000$ & $1.000 \pm 0.000$ \\
 & exact (control) & $3.3\times10^{-3}$ & $0.987 \pm 0.006$ & $0.025 \pm 0.003$ & -- \\
\midrule
\multirow{3}{*}{FNO (non-local)}
 & upwind & $2.9\times10^{-8}$ & $0.395 \pm 0.000$ & $0.004 \pm 0.001$ & $1.000 \pm 0.000$ \\
 & Crank--Nicolson & $6.5\times10^{-9}$ & $1.000 \pm 0.000$ & $0.825 \pm 0.000$ & $1.001 \pm 0.000$ \\
 & exact (control) & $3.6\times10^{-8}$ & $0.999 \pm 0.000$ & $0.001 \pm 0.000$ & -- \\
\midrule
\multicolumn{2}{l}{\emph{scheme value} (upwind / CN)} & -- & 0.395 / 1.000 & 0.000 / 0.824 & 1 \\
\bottomrule
\end{tabular}
\label{tab:arch}
\end{table}

Table~\ref{tab:arch} rules out the tested locality/capacity explanation. The FNO reproduces upwind's
amplitude deficit and Crank--Nicolson's phase error with provenance coefficients
statistically indistinguishable from the CNN's, and its twin difference reaches $99.9\%$
radial and $99.9\%$ tangential of the ceiling against the CNN's $100.0\%$ and $99.9\%$
(Appendix~\ref{app:twin_table}). At the same time the exact-trained control separates the two
architectures where the locality account says it should: trained on exact data the CNN cannot
fit the non-local evolution (MSE $3.3\times10^{-3}$), while the FNO fits it easily (MSE
$3.6\times10^{-8}$). The observed numerical provenance therefore cannot be explained solely by
insufficient capacity or CNN locality in this experiment; an architecture that represents the exact evolution when shown
exact data reproduces the scheme when shown solver data.

The FNO's exact-trained run also serves as a positive control for the instrument. A
diagnostic that returned ``scheme'' for every model would carry no information. Trained on
exact data, the FNO's symbol sits at the origin of the $(\Ddiss, \Ddisp)$ plane
($|\Ghat| = 0.999 \pm 0.000$ against a true value of $1$, mean lag $0.001 \pm 0.000$
radians), which realizes the ``consistent with exact reference'' row of
Table~\ref{tab:predictions} in a trained network. The same probe on the same band returns
$\alpha \approx 1$ when the training target is a scheme and $\approx 0$ when it is the exact
evolution; what changes is the training data, not the measurement.

\subsection{Nonlinear Burgers: provenance beyond linear dynamics}
\label{sec:burgers}

Linear advection admits an exact symbol for every operator involved, which is what makes it
the cleanest setting; the question here is whether numerical provenance survives when the
dynamics are nonlinear and the one-step map is not a Fourier multiplier. On forced viscous
Burgers the three solvers show the expected stationary-spectrum separation: the fraction of
energy above $k = N/4$ is $8.2\times10^{-7}$ for upwind, $3.3\times10^{-6}$ for the physical
spectral solver and $1.3\times10^{-5}$ for centered. There is therefore a scheme signature
present to be imitated.

We evaluate the surrogate's one-step \emph{operator} action on the solver's own forced steady
states, with no rollout, and compare its output spectrum both to the solver's true next-state
spectrum ($\Dscheme$) and to the dealiased spectral reference applied to the same inputs ($\Dexact$),
each measured as the mean $|\Delta \log E(k)|$ over the upper spectrum.

\begin{table}[t]\centering\small
\caption{Forced Burgers, one-step operator fingerprint evaluated on each surrogate's own
training-scheme steady states. $\Dscheme$ is the spectral distance between the surrogate's
output and its own solver's next state; $\Dexact$ is the distance between the surrogate's
output and the dealiased spectral reference operator applied to the same inputs; ``scheme
artifact'' is the corresponding distance between the two reference operators, i.e.\ the size
of the discretization error available to be imitated. The learned operator is $5$--$27\times$
closer to its training scheme than to the dealiased spectral reference.}
\begin{tabular}{lcccc}
\toprule
surrogate & $\Dscheme$ (vs.\ own solver) & $\Dexact$ (vs.\ spectral ref.) & ratio & scheme artifact \\
\midrule
upwind   & 0.564 & 3.02 & $5.4\times$ & 3.48 \\
centered & 0.127 & 3.40 & $26.8\times$ & 3.46 \\
\bottomrule
\end{tabular}
\label{tab:burgers}
\end{table}

Table~\ref{tab:burgers} shows the same preference found in the linear case, on an operator
that is not a linear stencil: each surrogate reproduces its scheme's spectral action while
diverging from the dealiased spectral reference, and the discretization error it could imitate is large
relative to that agreement.

We also repeat the comparison with a common input distribution, so that only the operator varies
(Appendix~\ref{app:burgers_common}). The learned operator remains substantially closer to its
training scheme's action than to the dealiased spectral reference action: $50.2\times$ for upwind and
Two additional checks clarify the scope of the diagnostic. First, autoregressive Burgers
rollout adds a scheme-independent over-smoothing that dominates the attractor-level spectrum,
so provenance must be measured at the operator level (Appendix~\ref{app:rollout}). Second,
fine-tuning a scheme-trained CNN on exact-evolution samples moves $\alpha$ from $1$ to
approximately $0$ with as few as $25$ samples, while exact-reference MSE remains at the
architecture-dependent floor (Appendix~\ref{app:migration}). Thus provenance and predictive
error are distinct quantities that can respond on different scales.

\section{Conclusion}

Solver-generated training data carry numerical provenance: the discretization error of the
scheme that produced them. Conventionally trained surrogates can reproduce that provenance,
so accuracy measured against solver-generated test sets does not by itself establish fidelity
to the exact evolution. The empirical Fourier symbol makes this provenance measurable, and
twin-scheme differencing makes the attribution robust to architectural spectral bias. The
signature is present for a local CNN, for a non-local FNO that can represent the exact
evolution, and at the operator level for nonlinear Burgers. Provenance and benchmark accuracy
are distinct quantities that can respond on different scales.

\bibliographystyle{plainnat}
\bibliography{references}

\appendix

\section{Measurement schematic}
\label{app:pipeline}

\begin{figure}[ht]\centering
\begin{tikzpicture}[
  font=\scriptsize,
  box/.style={draw, rounded corners=2pt, align=center, minimum height=8mm, inner sep=3pt},
  ar/.style={-{Latex[length=2mm]}, thick},
]
\node[box] (probe) {probe mode\\$\varepsilon\cos(2\pi k x/L)$};
\node[box, right=6mm of probe] (net) {one-step map\\$f_\theta$};
\node[box, right=6mm of net] (fft) {DFT, read\\bin $k$};
\node[box, right=6mm of fft] (sym) {symbol\\$\Ghat(k)$};
\node[box, right=7mm of sym] (proj) {rotate by $\overline{\Gex}$\\
  $(\Ddiss,\Ddisp)$};
\node[box, above right=4mm and 7mm of proj] (alpha) {$\alpha(k)$ against $\Gsch$\\
  \emph{provenance}};
\node[box, below right=4mm and 7mm of proj] (twin) {$\Ghat^{\mathrm{up}}-\Ghat^{\mathrm{CN}}$\\
  \emph{common $B(k)$ suppressed}};
\draw[ar] (probe) -- (net);
\draw[ar] (net) -- (fft);
\draw[ar] (fft) -- (sym);
\draw[ar] (sym) -- (proj);
\draw[ar] (proj) -- (alpha);
\draw[ar] (proj) -- (twin);
\node[below=1mm of net, font=\scriptsize\itshape, text width=26mm, align=center]
  {repeat per scheme};
\end{tikzpicture}
\caption{The measurement pipeline. One forward pass per wavenumber returns the empirical
symbol $\Ghat(k)$ of the trained surrogate's linearized one-step map (Sec.~\ref{sec:symbol}).
Multiplying the residual $\Ghat-\Gex$ by the complex conjugate $\overline{\Gex}$ rotates it
into the frame of the exact reference, whose real and imaginary parts are the dissipation and
dispersion coordinates (Sec.~\ref{sec:coords}). Projecting onto the direction
$\Gsch-\Gex$ gives the provenance coefficient $\alpha(k)$ (Sec.~\ref{sec:alpha});
differencing the symbols learned from two schemes suppresses the common component of the
architecture-intrinsic term $B(k)$ (Sec.~\ref{sec:twin}).}
\label{fig:pipeline}
\end{figure}
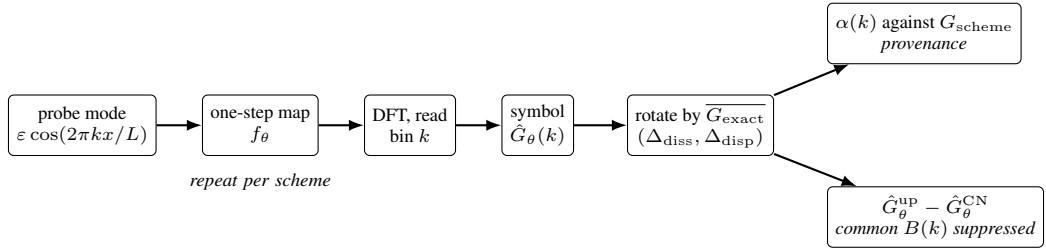

\section{Observable predictions}
\label{app:predictions}

\begin{table}[ht]\centering\small
\caption{What each account of surrogate error predicts, and what is observed. Columns are the
provenance coefficients of the two surrogates (\eqref{eq:alpha}) and the two components of
the twin-scheme difference (\eqref{eq:twin}); the last column asks whether the two learned
symbols separate along orthogonal directions in the $(\Ddiss,\Ddisp)$ plane. The
full-imitation ceiling is computed from the analytic symbols on this grid; the observed row
is the linear-advection CNN experiment of Sec.~\ref{sec:linear}, mean $\pm$ s.d.\ over
3 seeds. Only scheme imitation predicts a nonzero tangential twin difference.}
\setlength{\tabcolsep}{3.5pt}
\begin{tabular}{lccccc}
\toprule
account & $\alpha_{\mathrm{up}}$ & $\alpha_{\mathrm{CN}}$ & $\mathrm{Re}\,\Dtwin$ & $\mathrm{Im}\,\Dtwin$ & right angle? \\
\midrule
exact-reference-consistent & $0$ & $0$ & $0$ & $0$ & no (both at origin) \\
spectral bias only & $>0$ \emph{(spurious)} & $\approx0$ & $0$ & $0$ & no (both radial) \\
scheme imitation & $1$ & $1$ & $<0$ & $<0$ & \textbf{yes} \\
emulator superiority & $<0$ & $<0$ & $>0$ & $>0$ & no (past exact reference) \\
\midrule
\emph{full-imitation ceiling} & $1$ & $1$ & $-0.241$ & $-0.685$ & -- \\
\textbf{observed} & $1.00 \pm 0.00$ & $1.00 \pm 0.00$ & $-0.241$ & $-0.684$ & \textbf{yes} \\
\bottomrule
\end{tabular}
\label{tab:predictions}
\end{table}

\section{Twin-scheme architecture comparison}
\label{app:twin_table}

\begin{table}[ht]\centering\small
\caption{Twin-scheme difference $\Dtwin = (\Ghat^{\mathrm{up}} - \Ghat^{\mathrm{CN}})
\overline{\Gex}$ on linear advection, averaged over the upper spectrum, mean $\pm$ s.d.\ over
3 seeds, against the ceiling it would reach under full scheme imitation
($\Ghat^{(s)} = \Gsch^{(s)}$, at which the learned symbols have no residual $B(k)$). Both architectures
saturate both axes; the tangential axis, which a scheme-independent radial $B(k)$ cannot
produce, is as saturated as the radial one. The two architectures have very different
architectural displacements (their exact-trained controls differ by five orders of magnitude
in MSE, Table~\ref{tab:arch}) yet agree here to $4\times10^{-4}$, as expected if what
survives the difference is the scheme term.}
\begin{tabular}{lcccc}
\toprule
& radial (dissipation) & \% of ceiling & tangential (dispersion) & \% of ceiling \\
\midrule
CNN (local)     & $-0.2406 \pm 0.0001$ & $100.0\%$ & $-0.6841 \pm 0.0008$ & $99.9\%$ \\
FNO (non-local) & $-0.2402 \pm 0.0001$ & $99.9\%$ & $-0.6837 \pm 0.0003$ & $99.9\%$ \\
\midrule
\emph{full-imitation ceiling} & $-0.2406$ & -- & $-0.6845$ & -- \\
\bottomrule
\end{tabular}
\label{tab:twin}
\end{table}

\section{FNO representation of the advection step}
\label{app:fno_repr}

The capacity claim used in Sec.~\ref{sec:fno} is constructive at the level of the spectral
convolution. On a periodic grid, one exact step of $u_t + c u_x = 0$ is diagonal in the
Fourier basis:
$\widehat{u}_{k}(t+\Delta t)=e^{-i2\pi k c\Delta t/L}\widehat{u}_{k}(t)$. An FNO spectral
convolution that retains all resolved modes can assign this complex multiplier to each
Fourier coefficient. In a real-valued implementation, the complex multiplication is the
corresponding $2\times2$ rotation/scaling block on the real and imaginary parts of each mode.
For the full implemented network, which also contains pointwise maps and nonlinearities, this
argument is a capacity guide rather than a formal exact-representation theorem; the
exact-trained run in Table~\ref{tab:arch} is therefore the operative control, showing that
optimization finds an essentially exact symbol when the FNO is trained on exact data.

\section{Additional discussion and limitations}
\label{app:limitations}

The diagnostic measures relative alignment with two references, the exact evolution and the
training scheme. It does not decide whether a network has ``learned physics'' in an absolute
sense: $\alpha \approx 0$ only says that the scheme's error direction is absent from the
measured symbol. The evidence here covers one-dimensional periodic advection and forced
Burgers at modest resolution, two architectures and three seeds. For Burgers, the
``exact'' comparator is a dealiased spectral numerical reference, not a continuum oracle.

The identification argument requires $B(k)$ to be scheme-independent, and it is only
approximately so: the two corpora have different spectral content, so the optimization
trajectories are not identical. This is why we rely on out-of-distribution single-mode
probing and on differencing rather than on subtracting an exact-trained control, whose $B(k)$
is target-dependent. The architecture control gives some direct evidence that the
cancellation works: the CNN and the FNO have very different architectural displacements, since
their exact-trained controls differ by five orders of magnitude in MSE (Table~\ref{tab:arch}),
yet their twin differences agree to $4\times10^{-4}$ on both axes (Table~\ref{tab:twin}), which
is what one expects if the quantity that survives the difference is the scheme term.

The linear result is clean in part because a convolutional network represents a local stencil
almost exactly, and that is a genuine mechanism rather than an artifact: local architectures
carry an inductive bias toward discretization-like operators. The FNO control shows that this
mechanism is not the explanation, since removing locality does not remove the effect, and the
FNO's exact-trained run shows that the exact evolution was inside its hypothesis class the
whole time.

On Burgers we measure the learned one-step operator, linearized about states drawn from a
stated distribution, and we control the input distribution explicitly. We do not claim a
per-mode von~Neumann correspondence there, and the $\alpha$ values above unity on
off-distribution pools should be read as extrapolation along the scheme direction rather than
as a sharper measurement of imitation.
The dealiased spectral Burgers solver is used as a high-fidelity numerical reference; it is
not an exact continuum oracle.

Autoregressive rollout adds a scheme-independent over-smoothing that compounds with step
count and dominates the attractor-level spectrum (Appendix~\ref{app:rollout}), consistent
with reported long-rollout failures \citep{pderefiner}. It is a distinct failure mode with
its own literature; our claim about numerical provenance is a claim about the learned
operator.

An emulator can in principle be more accurate than the solver that trained it
\citep{emulatorsuperiority}, which in our coordinates means $\alpha < 0$. Our instrument
reports where a given model lands, and under standard training it lands at $\alpha \approx 1$.
We do observe the negative region once: the Crank--Nicolson-trained model fine-tuned on the
smallest budgets sits at $\alpha = -0.021$ to $-0.027$ (Table~\ref{tab:migration}),
consistently across seeds but at a magnitude close to our resolution. We report it as an
observation rather than as evidence for emulator superiority.

For one-step and short-horizon protocols scored against held-out solver output, the argument
is direct: the learned operator carries the scheme's symbol, the test targets carry the same
symbol, and the score rewards the agreement, so a high score does not establish fidelity to
the exact evolution. For long-rollout protocols the scheme signature is not what dominates; the
surrogate's own over-smoothing does. In both cases a solver-referenced score does not
decompose into a physical component and a numerical one, whereas the symbol does. Which term
dominates depends on the horizon, which is why the two are reported separately.

\section{Related work}
\label{app:related}

Von~Neumann analysis gives, for a linear scheme on a periodic grid, the exact per-mode
amplification factor $\Gsch(k)$, and modified-equation analysis identifies which
differential operator the scheme's leading error imitates: even-derivative terms produce
amplitude error (dissipation), odd-derivative terms produce phase error (dispersion)
\citep{leveque}. Fourier pseudospectral methods and aliasing control, including the
two-thirds rule used in our Burgers reference solver, are standard numerical-analysis tools
\citep{boyd}. This literature supplies the reference objects we attribute against. It does
not, by itself, identify what a learned operator has inherited from its training solver.

Neural operators such as the FNO learn solution operators from solver-generated data
\citep{fno}; learned simulators and neural PDE solvers are commonly trained and evaluated by
autoregressive comparison to numerical rollouts \citep{gns,mppde}. Benchmarks such as
PDEBench and PDEArena standardize this practice across datasets and architectures
\citep{pdebench,pdearena}. These protocols make large-scale comparison possible, but their
targets are numerical outputs, not separate measurements of exact-evolution fidelity and
scheme fidelity.

Neural networks fit low frequencies first and represent high frequencies poorly
\citep{rahaman,tancik}, and the effect persists in physics-informed and operator learning
\citep{spectralbiasop,pderefiner}. This is a property of the architecture and optimization rather than
of the training targets, which is precisely what makes it a confound here: it displaces a
learned symbol in the same direction a dissipative scheme does. Our design treats it as a
nuisance term to be cancelled rather than as the effect under study.

Suriyanarayanan, Chakraborty, and Maulik \citep{solverclosures} show that learned turbulence closures inherit the discretization error
of the solver used to generate their training data, and condition the model on the solver to
exploit this. The evidence there is aggregate: energy spectra that differ by scheme, without
a per-mode amplification factor, without separating amplitude from phase error, and without a
scheme-free reference operator to attribute against. Related work learns correction terms
inside differentiable solvers \citep{solverintheloop}, learns stencil-like differential
operators \citep{pdenet}, or combines low- and high-fidelity operator data
\citep{multifidelitydeeponet}. Koehler and Thuerey analyze theoretically when an
emulator can be more accurate than the low-fidelity solver that trained it, the regime we
label $\alpha < 0$, but assess trained networks through aggregate error and rollout rather
than through their learned symbols. Our question is complementary: after ordinary supervised
training on a fixed solver corpus, where does the learned one-step operator lie relative to
the exact reference and to the training scheme?

\citet{spectralaudit} extract the per-mode gain and phase of a trained operator network's
Jacobian and compare them to the exact advection symbol. We adopt that measurement primitive.
The difference is what the symbol is compared against: they compare to the exact operator
only, and so measure error, whereas we compare to the exact operator \emph{and} to the
training scheme, label the two error directions, and difference two schemes to remove the
architectural contribution. Among the closest diagnostics and solver-aware methods above, we
are not aware of one that attributes a learned operator to a specific training discretization
by comparing its Fourier response to both the exact and scheme references.

Long autoregressive rollouts of neural PDE surrogates degrade in characteristic ways, and
correcting that degradation is an active area \citep{gns,mppde,pderefiner}. Noise injection,
pushforward training and refinement-based methods all target the distribution shift and
frequency loss that appear when a one-step model is used autoregressively. Our rollout
measurements are consistent with this literature and are the reason we report provenance on
the one-step operator: rollout introduces a distortion that belongs to the surrogate rather
than to either reference operator, and it can dominate the attractor-level spectrum.

\section{Additional linear-advection checks}
\label{app:linear_extra}

Deliberately restricting the surrogate (small width, few epochs, red-spectrum training data)
to induce a larger architectural displacement leaves the twin-scheme difference essentially
unchanged (radial $-0.240$, tangential $-0.686$), whereas subtracting an exact-trained control
is fragile because $B(k)$ depends on the training target. Twin differencing is therefore the
estimator we rely on. The restricted scheme-trained surrogates themselves remain faithful to
their schemes, because upwind and Crank--Nicolson are local finite-stencil operators that a
convolutional network represents easily \citep{pdenet,mppde}. The exact-trained control is the run that degrades,
and it degrades at full capacity too, plateauing four orders of magnitude above the scheme
corpora (Table~\ref{tab:arch}).

\section{Burgers common-input control}
\label{app:burgers_common}

For the common-input control, on a common pool $X$ we apply the surrogate, its own scheme's
$M$-step operator, and the dealiased spectral reference operator to the same inputs. Because the inputs coincide,
the outputs are comparable pointwise, and we report the same projection used in the linear
case,
\begin{equation}
  \alpha \;=\;
  \frac{\big\langle Y_{\mathrm{surr}} - Y_{\mathrm{spec}},\;
        Y_{\mathrm{solv}} - Y_{\mathrm{spec}}\big\rangle}
       {\big\|Y_{\mathrm{solv}} - Y_{\mathrm{spec}}\big\|^2},
  \label{eq:alpha_field}
\end{equation}
now in field space rather than per mode. A projection of this form cannot be inflated merely
because the input pool makes every operator move a great deal.

\begin{table}[ht]\centering\small
\caption{Forced Burgers, one-step operator with the input distribution held fixed and only
the operator varying. $\alpha$ is the field-space projection \eqref{eq:alpha_field}, with
$1$ meaning the learned operator reproduces its training scheme and $0$ meaning it reproduces
the dealiased spectral reference; brackets are $95\%$ bootstrap confidence intervals over trajectories.
$\Dscheme$, $\Dexact$, ratio and artifact are as in Table~\ref{tab:burgers}, now computed on
the common pool. The first block recomputes the protocol of Table~\ref{tab:burgers}; the
lower blocks evaluate every operator on spectral-reference states and on a mixture of all three
solvers' states. Fixing the input distribution does not remove the preference for the
training scheme.}
\begin{tabular}{llccccc}
\toprule
input pool & surrogate & $\alpha$ & $\Dscheme$ & $\Dexact$ & ratio & artifact \\
\midrule
\multirow{2}{*}{own scheme's states} & upwind   & $1.032$ [1.023, 1.040] & 0.564 & 3.02 & $5.4\times$ & 3.48 \\
                                     & centered & $0.946$ [0.937, 0.954] & 0.127 & 3.40 & $26.8\times$ & 3.46 \\
\midrule
\multirow{2}{*}{spectral-reference states} & upwind   & $2.845$ [2.827, 2.867] & 0.754 & 37.87 & $50.2\times$ & 38.11 \\
                                            & centered & $2.835$ [2.777, 2.906] & 0.720 & 37.72 & $52.4\times$ & 37.00 \\
\midrule
\multirow{2}{*}{mixed (all three)} & upwind   & $2.836$ [2.782, 2.896] & 0.987 & 3.50 & $3.5\times$ & 3.87 \\
                                   & centered & $1.968$ [1.869, 2.070] & 0.142 & 3.46 & $24.4\times$ & 3.43 \\
\bottomrule
\end{tabular}
\label{tab:burgers_common}
\end{table}

\section{Rollout: a second, scheme-independent distortion}
\label{app:rollout}

The measurements in the main paper are all one-step. The question here is whether the same
signature is visible when the surrogate is run autoregressively to its own statistical steady
state, which is how surrogates are often assessed.

\begin{figure}[ht]\centering
\includegraphics[width=\linewidth]{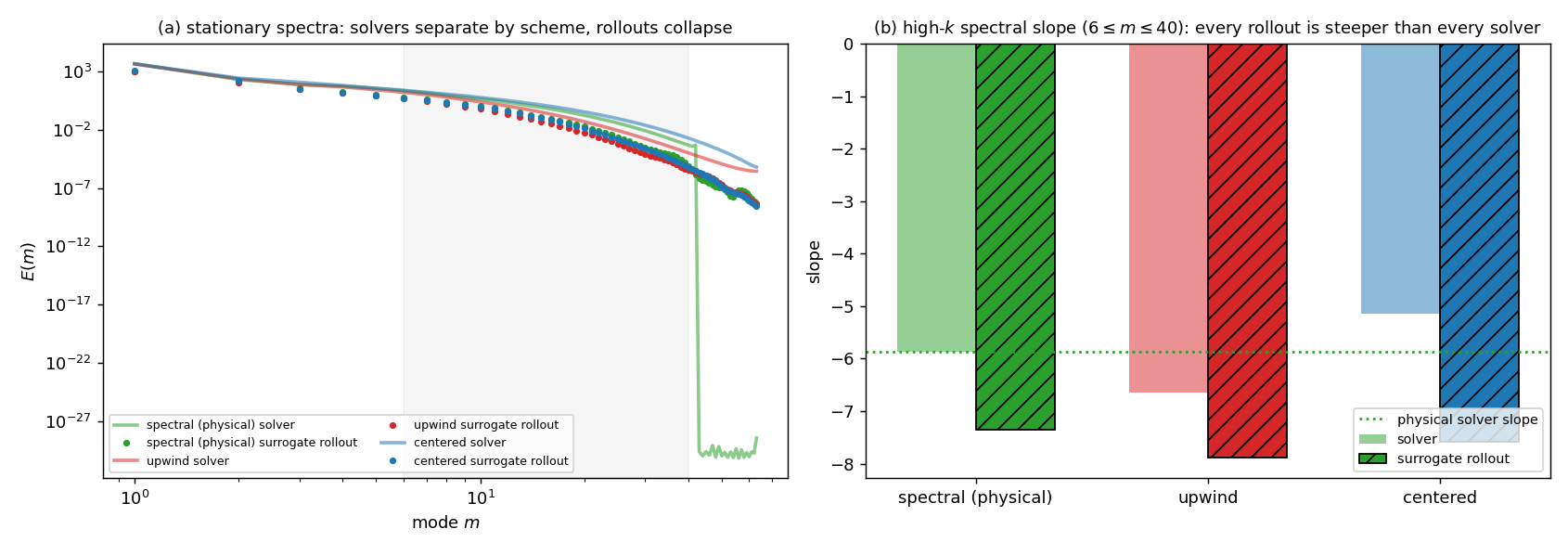}
\caption{Forced Burgers: operator-level provenance is not visible at the attractor.
\textbf{(a)} stationary energy spectra $E(m)$ against mode $m$ (log--log). Solid lines are the
three solvers, which separate by scheme as modified-equation theory predicts (centered above
the spectral reference above upwind at high $m$); dots are the corresponding surrogates run
autoregressively to their own steady state, which collapse onto one another and fall below
every solver. The shaded band is the fitting range for panel (b). \textbf{(b)} high-$k$
spectral slope fitted over $6 \leq m \leq 40$: the solvers straddle the spectral-reference slope
(dotted line), while every surrogate rollout is steeper than every solver, i.e.\
autoregressive over-smoothing dominates the scheme differences. The scheme signature these
same models carry in their one-step operators (Tables~\ref{tab:burgers}
and~\ref{tab:burgers_common}) is therefore not recoverable from rollout statistics.}
\label{fig:burgers}
\end{figure}

It is not. Rolled out autoregressively, all three Burgers surrogates over-damp: their
high-$k$ spectral slopes fall to between $-7.4$ and $-7.9$ against solver slopes of $-5.1$ to
$-6.7$, their high-$k$ energy lies below that of every solver, and the ordering by scheme
visible in the solvers is lost (Figure~\ref{fig:burgers}). This over-smoothing is
scheme-independent and compounds with the number of steps, and at the attractor it dominates
the scheme signature that Sec.~\ref{sec:burgers} finds in the same models' one-step
operators.

The scope of the diagnostic follows from this. Numerical provenance is a property of the
learned operator and must be measured there; an attractor-level spectrum mixes it with a
distortion that belongs to the surrogate's autoregressive use rather than to either reference
operator. This is also why we report the operator and the rollout separately rather than
combining them into a single score.

\section{High-accuracy fine-tuning: provenance and accuracy respond on different scales}
\label{app:migration}

The practical question the diagnostic raises is what it would take to move a surrogate's
provenance off its training scheme. We train to convergence on a scheme corpus, so that
$\alpha \approx 1$ at the start, then fine-tune on a budget of exact-evolution samples,
sweeping the budget, and re-measure the symbol. Because $\alpha$ is read from the symbol
rather than from the loss, it reports which of the two references the model is converging to,
which the loss curve alone does not.

\begin{table}[ht]\centering\small
\caption{Provenance under high-accuracy fine-tuning (mean $\pm$ s.d., 3 seeds). Budget is the
number of exact-evolution samples used for fine-tuning, against an original scheme corpus of
$4000$ samples. $\alpha$ is measured against the \emph{original} training scheme: $1$ means
still scheme-consistent, $0$ means the scheme's error direction is no longer present.
MSE$_{\mathrm{exact}}$ and MSE$_{\mathrm{scheme}}$ are one-step errors against the exact
evolution and against the original training scheme; together they show that the model does
not interpolate between the two operators but changes target, since error against the exact
evolution falls while error against the original scheme rises by six orders of magnitude.
Accuracy against the exact evolution saturates at $3.5\times10^{-3}$, the same floor the
exact-trained CNN reaches in Table~\ref{tab:arch}.}
\setlength{\tabcolsep}{3.5pt}
\begin{tabular}{lcccccc}
\toprule
& \multicolumn{3}{c}{upwind-trained} & \multicolumn{3}{c}{CN-trained} \\
\cmidrule(lr){2-4}\cmidrule(lr){5-7}
budget & $\alpha$ & MSE$_{\mathrm{exact}}$ & MSE$_{\mathrm{scheme}}$ & $\alpha$ & MSE$_{\mathrm{exact}}$ & MSE$_{\mathrm{scheme}}$ \\
\midrule
0 & $1.000 \pm 0.000$ & $1.1\times10^{-1}$ & $1.4\times10^{-7}$ & $1.000 \pm 0.000$ & $1.8\times10^{-1}$ & $4.1\times10^{-8}$ \\
25 & $0.008 \pm 0.006$ & $5.7\times10^{-3}$ & $1.0\times10^{-1}$ & $-0.021 \pm 0.010$ & $6.6\times10^{-3}$ & $1.8\times10^{-1}$ \\
50 & $0.006 \pm 0.001$ & $5.6\times10^{-3}$ & $1.0\times10^{-1}$ & $-0.026 \pm 0.005$ & $6.5\times10^{-3}$ & $1.8\times10^{-1}$ \\
100 & $0.007 \pm 0.005$ & $5.6\times10^{-3}$ & $1.0\times10^{-1}$ & $-0.027 \pm 0.006$ & $6.5\times10^{-3}$ & $1.8\times10^{-1}$ \\
250 & $0.013 \pm 0.000$ & $4.4\times10^{-3}$ & $1.0\times10^{-1}$ & $0.003 \pm 0.005$ & $4.9\times10^{-3}$ & $1.8\times10^{-1}$ \\
500 & $0.005 \pm 0.003$ & $3.6\times10^{-3}$ & $1.0\times10^{-1}$ & $0.007 \pm 0.001$ & $3.9\times10^{-3}$ & $1.8\times10^{-1}$ \\
1000 & $0.007 \pm 0.004$ & $3.5\times10^{-3}$ & $1.0\times10^{-1}$ & $0.000 \pm 0.002$ & $3.5\times10^{-3}$ & $1.8\times10^{-1}$ \\
4000 & $0.007 \pm 0.002$ & $3.5\times10^{-3}$ & $1.0\times10^{-1}$ & $-0.001 \pm 0.000$ & $3.5\times10^{-3}$ & $1.8\times10^{-1}$ \\
\bottomrule
\end{tabular}
\label{tab:migration}
\end{table}

Table~\ref{tab:migration} shows two things that move differently. Provenance changes at the
smallest budget tested: $25$ exact samples, under one percent of the original training set,
take $\alpha$ from $1.000$ to $0.008$ for the upwind-trained model and $-0.021$ for the
Crank--Nicolson-trained model, and every larger budget leaves it there. The change is a
change of target rather than a blend, since error against the original scheme rises from
$1.4\times10^{-7}$ to $1.0\times10^{-1}$ over the same step.

Accuracy does not follow. Error against the exact evolution improves but saturates at
$3.5\times10^{-3}$, which is the floor the exact-trained CNN reaches in Table~\ref{tab:arch}
and is therefore a property of the architecture rather than of the data budget. The two
coordinates come apart: $\alpha$ reports that the surrogate no longer carries its solver's
error direction, while the loss reports that it still cannot represent the exact operator.
Neither implies the other. We state this as a result about this setting, in which a small
high-fidelity budget changed provenance and the loss curve alone would not have shown it,
rather than as a general claim that $25$ samples remove inherited numerics.

\section{Training details and the identical-conditions requirement}
\label{app:details}

The twin-scheme argument requires $B(k)$ to be scheme-independent, and $B(k)$ is a property
of the architecture and of the optimization. Everything below is therefore held identical
across the schemes within an experiment: architecture, initialization seed sequence,
optimizer, learning-rate schedule, batch size, epoch count, sample count, and the
distribution of input fields. The corpora differ only in the solver that generated the
targets.

$u_t + u_x = 0$ on $[0,1]$, $N = 64$, $\sigma = c\Delta t/\Delta x = 0.5$. Training inputs are
random band-limited fields with approximately flat spectra,
$u(x) = \sum_{m=1}^{N/2-1} a_m \cos(2\pi m x + \phi_m)$ with
$a_m \sim \mathcal{N}(0, 1/\sqrt{N/2})$ and $\phi_m \sim \mathcal{U}[0,2\pi)$; targets are one
solver step. $4000$ samples, no validation split, since the symbol probe is the evaluation.
\emph{CNN:} five periodic \texttt{Conv1d} layers, kernel $5$, width $64$, GELU \citep{gelu}, residual
output $u + \mathrm{net}(u)$; receptive field $\pm 10$ cells. \emph{FNO:} $1\times1$ lifting
to width $32$, four spectral-convolution blocks retaining all $N/2$ resolved modes
\citep{fno} with a
$1\times1$ bypass, GELU, two-layer $1\times1$ projection, same residual form. Both: Adam
\citep{adam},
$\eta_0 = 2\times10^{-3}$, cosine annealing \citep{sgdr} over $400$ epochs, batch $128$, MSE loss. Three
seeds ($0$--$2$); all reported figures are mean $\pm$ s.d.\ over seeds. Symbol statistics are
averaged over the upper half of the resolved spectrum, $k \geq N/4$.

$u_t + u u_x = \nu u_{xx} + f$ on $[0,2\pi]$, $N = 128$, $\nu = 0.04$,
$\Delta t = 1.5\times10^{-3}$, steady forcing
$f = 0.3(\sin x + \sin(2x+1) + \tfrac12 \sin(3x+2))$. All three solvers are method-of-lines
with RK4 and share mesh, $\Delta t$, $\nu$ and $f$, differing only in the convection operator
(dealiased spectral with the $2/3$ rule \citep{boyd}; first-order upwind; second-order centered
\citep{leveque}). The
surrogate learns the $M = 15$ substep map. Data: $16$ trajectories, $2500$ warmup steps, then
$200$ collected states, formed into $(t, t+M)$ pairs taken from the stable trajectory rather
than by re-stepping sampled snapshots, which diverges for the centered scheme. \emph{Net:}
six periodic \texttt{Conv1d} layers, kernel $5$, width $96$, GELU, residual. Adam,
$\eta_0 = 5\times10^{-4}$, cosine annealing over $250$ epochs, batch $64$, gradient-norm clipping at
$1.0$, and input noise of $0.03\,\mathrm{std}(X)$ injected each step, which is standard
practice for autoregressive stability \citep{gns,mppde}.

Models trained to convergence on a scheme corpus are fine-tuned on budgets of
$25$--$4000$ exact-evolution samples drawn from the same input distribution, with the symbol
re-measured after fine-tuning and $\alpha$ evaluated against the original training scheme.
Three seeds throughout.

\section{Linearity of the symbol probe}
\label{app:eps}

Equation~\eqref{eq:symbol} defines a symbol only in the linear regime of the probe amplitude
$\varepsilon$. Table~\ref{tab:eps} reports the relative change of the recovered symbol as
$\varepsilon$ is swept over the tested range, relative to the value at the operating point
$\varepsilon = 10^{-2}$. The worst case over all models and amplitudes in
$[10^{-4}, 10^{-1}]$ is $1.0\times10^{-2}$, and the probe departs from linearity only as
$\varepsilon \to 1$, where the GELU nonlinearity is no longer probed locally.

\begin{table}[ht]\centering\small
\caption{Linearity of the symbol probe:
$\max_k |\Ghat(\varepsilon) - \Ghat(10^{-2})| / |\Ghat(10^{-2})|$ over the resolved spectrum,
for each architecture/corpus combination. The column at $10^{-2}$ is the reference point. The
recovered symbol is stable over the tested $10^{-4}$--$10^{-1}$ range, with worst-case relative
change $1\%$.}
\begin{tabular}{lcccc}
\toprule
model & $10^{-4}$ & $10^{-3}$ & $10^{-2}$ & $10^{-1}$ \\
\midrule
\texttt{cnn/cn} & $5.7\times10^{-5}$ & $7.3\times10^{-6}$ & -- & $7.1\times10^{-6}$ \\
\texttt{cnn/exact} & $1.8\times10^{-4}$ & $10.0\times10^{-5}$ & -- & $1.0\times10^{-2}$ \\
\texttt{cnn/upwind} & $3.4\times10^{-4}$ & $5.0\times10^{-5}$ & -- & $2.4\times10^{-3}$ \\
\texttt{fno/cn} & $6.2\times10^{-5}$ & $4.8\times10^{-6}$ & -- & $2.5\times10^{-5}$ \\
\texttt{fno/exact} & $9.1\times10^{-5}$ & $1.2\times10^{-5}$ & -- & $7.7\times10^{-5}$ \\
\texttt{fno/upwind} & $9.7\times10^{-4}$ & $1.2\times10^{-4}$ & -- & $7.5\times10^{-4}$ \\
\bottomrule
\end{tabular}
\label{tab:eps}
\end{table}

\section{Numerical reproducibility notes}
\label{app:repro}

Two implementation issues produced silently incorrect results during this work and are
recorded here because both yield plausible output rather than an error.

Our first FNO runs did not train, stalling just below the loss of the identity map. The cause
was not the model but the backward pass through a complex-valued \texttt{einsum} on the MPS
backend, which we measured to be wrong by $\sim\!140\%$ relative error in
$\partial \mathcal{L}/\partial W$ against a CPU reference, while \texttt{rfft}/\texttt{irfft}
autograd agreed to $2\times10^{-7}$. Rewriting the mode-wise multiplication in real arithmetic
restores agreement with the CPU reference to $4\times10^{-7}$. The failure mode matters for
the conclusions of Sec.~\ref{sec:fno}: an untrained FNO can produce a conclusion that is
numerically plausible and scientifically incorrect, namely that non-local architectures do not
reproduce their training scheme.
Any new architecture here is checked against the identity-map loss before its symbols are
read.

Iterating RK4 over a large batch in a single array corrupted values when the total number of
elements in the batched state array exceeded $2^{15}$, either producing NaNs (with a
finite-difference right-hand side) or,
more dangerously, staying finite while drifting several percent (with a spectral right-hand
side). Integration is now chunked well below that threshold and verified at run time against
a per-row reference. The Burgers reference operators are recomputed under this fix; a field
error of a few percent had translated into a factor of about two in
mean $|\Delta \log E(k)|$ over high wavenumbers, so upstream perturbations of this kind cannot
be assumed small in the derived spectral statistics. All affected Burgers results reported in
this paper were regenerated after the correction and verified against the per-row reference.
The linear-advection results are unaffected, as those solvers are applied per sample.

\end{document}